\documentclass[referee,sn-basic]{sn-jnl} 

\usepackage{natbib}
\usepackage{graphicx}%
\usepackage{amsmath,amssymb,amsfonts}%
\usepackage{amsthm}%
\usepackage{mathrsfs}%
\usepackage[title]{appendix}%
\usepackage{xcolor}%
\usepackage{textcomp}%
\usepackage{manyfoot}%
\usepackage{algorithm}%
\usepackage{algorithmicx}%
\usepackage{algpseudocode}%
\usepackage{listings}%
\usepackage{multirow}%
\usepackage{array}%
\usepackage{comment}%

\DeclareFontFamily{U}{rsfs}{}
\DeclareFontShape{U}{rsfs}{m}{n}{<-> rsfs10}{}

\usepackage{multirow} 
\usepackage{booktabs} 
\usepackage{pdfpages}
\usepackage{pdflscape}

\usepackage{xcolor}

\usepackage{microtype}
\usepackage{fullpage}
\usepackage{tcolorbox}

\begin{document}

\title[Article Title]{PROMPTHEUS: A Human-Centered Pipeline to Streamline SLRs with LLMs}


\author[1]{\fnm{Joao Pedro Fernandes} \sur{Torres}}\email{joao.fernandes.torres@tecnico.ulisboa.pt}

\author[3]{\fnm{Catherine} \sur{Mulligan}}\email{c.mulligan@imperial.ac.uk}

\author[1]{\fnm{Joaquim} \sur{Jorge}}\email{jorgej@tecnico.ulisboa.pt}

\author*[2]{\fnm{Catarina} \sur{Moreira}}\email{catarina.pintomoreira@uts.edu.au}

\affil*[1]{\orgdiv{INESC-ID}, \orgname{DEI/IST/UL} \country{Portugal}}

\affil[2]{\orgdiv{Data Science Institute}, \orgname{University of Technology Sydney}, \orgaddress{
\city{Sydney},  \country{Australia}}}

\affil[3]{\orgdiv{ISST, Imperial College London}, \orgname{\city{London},  \country{UK}}}

\abstract{
The growing volume of academic publications poses significant challenges for researchers conducting timely and accurate Systematic Literature Reviews, particularly in fast-evolving fields like artificial intelligence. This growth of academic literature also makes it increasingly difficult for lay people to access scientific knowledge effectively, meaning academic literature is often misrepresented in the popular press and, more broadly, in society.  Traditional SLR methods are labor-intensive and error-prone, and they struggle to keep up with the rapid pace of new research. To address these issues, we developed \textit{PROMPTHEUS}: an AI-driven pipeline solution that automates the SLR process using Large Language Models. We aimed to enhance efficiency by reducing the manual workload while maintaining the precision and coherence required for comprehensive literature synthesis. PROMPTHEUS automates key stages of the SLR process, including systematic search, data extraction, topic modeling using BERTopic\footnote{https://maartengr.github.io/BERTopic/}, and summarization with transformer models. Evaluations conducted across five research domains demonstrate that PROMPTHEUS reduces review time, achieves high precision, and provides coherent topic organization, offering a scalable and effective solution for conducting literature reviews in an increasingly crowded research landscape. In addition, such tools may reduce the increasing mistrust in science by making summarization more accessible to laypeople.

The code for this project can be found on the GitHub repository at \url{https://github.com/joaopftorres/PROMPTHEUS.git}
}




\maketitle

\section{Introduction}\label{sec1}

The exponential growth of academic publications poses a significant challenge for researchers attempting to stay current with developments across numerous fields. Over 2.5 million papers are published annually across approximately 30,000 accredited journals worldwide, making it increasingly challenging to filter through relevant research efficiently. This is particularly evident in rapidly evolving domains such as artificial intelligence, virtual reality, and blockchain technologies \citep{STMreport}. As the research landscape becomes saturated, the visibility of new and impactful studies diminishes, complicating the synthesis of existing knowledge \citep{CanonicalProgress}. Systematic Literature Reviews (SLRs) provide a structured approach to summarizing and synthesizing research, offering essential insights into specific topics. However, traditional, manual SLR approaches are labor-intensive, prone to human error, and increasingly unsustainable due to this overwhelming data volume.

Despite advancements in AI-driven tools for literature searches and summarization, a significant gap remains in fully automating the systematic review process. While individual tools exist for searching, filtering, and summarizing content, they often operate in isolation, leaving researchers to handle other crucial stages manually, such as selecting relevant literature, extracting insights, and generating coherent reports. The scientific community lacks a fully integrated, scalable solution that automates the SLR pipeline. Such a solution must ensure accuracy and efficiency while reducing the manual burden on researchers.

In response to this need, we propose \textit{PROMPTHEUS: PRocess Optimization and Management of Papers using emerging Technologies for High Efficiency in Updated Systematic Reviews}. PROMPTHEUS is an automated framework that integrates Large Language Models (LLMs) to automate key phases of the SLR process: Systematic Search and Screening, Data Extraction, and Synthesis and Summarization. While the critical planning phase remains in the hands of researchers, PROMPTHEUS significantly reduces manual workload, enhancing the precision, accuracy, and relevance of final outputs, thus allowing researchers to focus more on the innovative aspects of their work.

The contributions of this work are the following:
\begin{itemize}
    \item \textbf{Novel Integration of SLR Phases:} We present a fully automated approach to SLRs, combining multiple stages—search, extraction, and synthesis—into an end-to-end process powered by advanced natural language processing (NLP) techniques.

    \item \textbf{Precision in Literature Retrieval:} We leverage state-of-the-art language models to enhance the precision of literature searches. This ensures that researchers receive high-quality and relevant studies, addressing a critical need for accurate literature filtering.

    \item \textbf{Structured Topic Modeling:} PROMPTHEUS employs BERTopic, a topic modeling technique that structures the extraction and organization of information, allowing for clear, well-organized reviews.

    \item  \textbf{Comprehensive Evaluation:} We present a robust evaluation using several metrics, including ROUGE scores, Flesch readability scores, cosine similarity, and topic coherence. These evaluations demonstrate the effectiveness of PROMPTHEUS in automating the SLR process while maintaining high accuracy and improving the readability of generated content.
\end{itemize}

By automating the most time-consuming aspects of systematic literature reviews, PROMPTHEUS aims to make SLRs more accessible, efficient, and comprehensive. This will ultimately enable researchers to devote more time to innovative, high-impact research while ensuring they remain up-to-date with critical developments.

\section{Background and Related Work}

Systematic Literature Reviews are crucial for synthesizing research, identifying knowledge gaps, and shaping future directions across various domains. The traditional SLR process, defined by the PRISMA guidelines \citep{Page2021, Page2021}, consists of four phases: Planning, Selection, Extraction, and Execution \citep{SLRguide}. Despite its rigor, this method faces increasing challenges due to the sheer volume of academic publications. The manual nature of SLRs makes them labor-intensive, prone to error, and difficult to scale, particularly as research outputs grow exponentially.

\subsection{Advances in Automating Systematic Literature Reviews}

Recently, machine learning (ML) and natural language processing have emerged as powerful tools that can assist with these challenges by automating various SLR process stages. Large language models such as T5 \citep{t5}, GPT-3.5, GPT-4o, and GPT-o1 \citep{gpt3} have been integrated into the SLR workflow, particularly for tasks like literature search, data extraction, and summarization. These AI approaches, including Technology-Assisted Review (TAR) systems, apply NLP and ML techniques to automate the search and screening phases, significantly reducing manual effort by iteratively refining models to prioritize relevant studies. This automation extends to data extraction, where NLP techniques ensure consistency and synthesis, where models such as T5 and GPT generate coherent summaries of research findings, enhancing accuracy and readability.

\citet{moreno2023novel} propose a novel AI-based framework that leverages ensemble learning techniques to improve the accuracy and efficiency of study selection and data extraction processes. Their model demonstrates how ensemble techniques, when applied to AI-assisted systematic reviews, can enhance the precision and recall of study identification while reducing manual effort. This work contributes to AI-driven SLR tools by highlighting the potential for combining multiple AI models to tackle the inherent variability and challenges in automating complex tasks like data extraction and synthesis.

\citet{bolanos2024artificial} conducted a comprehensive review of AI-integrated SLR tools, highlighting the efficiency improvements AI brings while emphasizing usability-related challenges. The authors highlight the need for user-friendly tools and strategies to manage LLM hallucinations, notably through knowledge injection techniques. Similarly, \citet{Saeidmehr2024spiral} proposed a spiral approach to systematic reviews, significantly improving screening efficiency in smaller datasets while also addressing gaps in handling unbalanced datasets and improving article acquisition.

Other AI-based models, such as the multi-agent AI system developed by \citet{sami2024}, offer a promising approach by automating most steps in the SLR process. Their system uses LLMs to automate tasks like generating search strings and screening abstracts. However, while this approach reduces the manual workload, it still faces limitations in managing complex queries and ensuring the relevance of the selected studies.

Automation techniques are increasingly used in systematic reviews, reducing manual workloads by up to 7\%, as noted by \citet{toth2024automation}. However, challenges remain in recall consistency and real-world adoption. The study emphasizes the need for standardized evaluation metrics to better assess automation's impact, showing that while promising, automation's full potential is not yet realized due to technical and practical limitations.

\subsection{Limitations of Current Automated SLR Systems}

Despite the progress in AI-assisted SLRs, several limitations remain. Many systems struggle with handling complex queries, often relying on simple keyword searches that fail to capture the depth and specificity needed for comprehensive reviews. Additionally, the criteria for inclusion and exclusion are frequently poorly defined, leading to the retention of irrelevant or low-quality studies. Existing research highlights the need for more sophisticated search algorithms, improved Boolean logic integration, and better strategies for managing large datasets without sacrificing accuracy and relevance \citep{chappell2023machine, guo2024rapid, robledo2023vista}.

Machine learning tools, such as \textit{Abstrackr}, have effectively automated SLRs' title and abstract screening stages. In the study by \citet{gates2020decoding}, Abstrackr reduced manual effort by up to 35\%, significantly saving time while maintaining a high accuracy level in identifying relevant studies. However, the tool missed some important studies during the screening process, underscoring a critical limitation of AI-assisted systems. Despite the efficiency gains, these tools still require human oversight to ensure that essential research is not inadvertently excluded. This highlights the balance between leveraging AI to reduce workload and ensuring that expert validation is in place to preserve the comprehensiveness and quality of the review process.

To support these findings, \citet{cierco2022machine} reviewed a range of ML tools for automating the SLR process, noting that many tools lacked user-friendly interfaces for researchers without programming skills. \citet{affengruber2024exploration} presented this finding, showing that while tools like \textit{Abstrackr} and \textit{Rayyan} can enhance efficiency, there is still a need for more comprehensive evaluations of their usability and impact in real-world scenarios. In addition, \citet{perlman2023real} evaluated an NLP tool for abstract screening during a SARS-CoV-2 review, reducing screening time by 33.7\% while still requiring human oversight to ensure accuracy.

While AI has improved screening efficiency, challenges remain in automating tasks like data extraction and risk of bias assessment. \citet{ofori2024towards} highlight that screening is the most automated phase, but more advanced AI techniques are needed for accurate data extraction and bias assessment. This shows that while automation reduces workloads, it still falls short in handling complex tasks in systematic reviews.

The role of human expertise remains critical in maintaining the rigor of systematic reviews, particularly when AI tools are not yet fully capable of handling the complexities of comprehensive research synthesis. \citet{qureshi2023chatgpt} and \citet{li2024evaluating} both highlighted the limitations of ChatGPT and similar LLMs, showing that while these models excel in specific tasks like abstract screening, they often require expert validation to prevent the inclusion of irrelevant or erroneous studies.  Rather than replacing human researchers, these tools are, therefore, best viewed as assistants in keeping up to date with emerging new work in the field.

\subsection{Advanced NLP and LLM Techniques}

Recent studies demonstrate the potential of advanced NLP techniques in addressing some of the limitations of current automated SLR systems. For instance, \citet{kharawala2021artificial} explored using zero-shot classification combined with ML algorithms to automate abstract screening, demonstrating high precision and recall. Similarly, \citet{dennstadt2024title} tested LLMs for title and abstract screening in the biomedical domain, showing high sensitivity but noting challenges related to resource demands and biases.

To enhance AI's role in systematic reviews, \citet{hamel2021guidance} developed a framework for integrating AI into the title and abstract screening phases of SLRs, stressing the importance of robust training sets and transparent reporting. In parallel, \citet{masoumi2024natural} demonstrated the effectiveness of BioBERT, a variant of BERT fine-tuned for biomedical texts, in automating the abstract review process in medical research, showing that such models can significantly reduce manual workloads while maintaining high accuracy.

\subsection{Challenges of Rapid Reviews and Methodological Shortcuts}

AI-based approaches have also been applied to rapid reviews (RRs), often employing methodological shortcuts to expedite the review process. \citet{guo2024rapid} examined the impact of these shortcuts, showing that while they improve efficiency, they can introduce biases and reduce comprehensiveness. \citet{speckemeier2022methodological} echoed these concerns, calling for more rigorous methodologies to balance the need for speed with the maintenance of review quality.

Moreover, \citet{o2019question} examined the cultural and practical challenges of adopting automation tools in systematic reviews, particularly in healthcare. Their study emphasized better collaboration between AI systems and human experts to ensure these tools are effectively integrated into existing workflows.

\subsection{Addressing the Gap}

Building on the limitations identified in existing automated systems, our work presents \textit{PROMPTHEUS}. This fully automated SLR pipeline system enhances the review process by addressing critical limitations such as inadequate inclusion/exclusion criteria and complex query handling. PROMPTHEUS automates the Selection, Extraction, and Synthesis phases, allowing researchers to manage the Planning phase while leveraging advanced NLP techniques like BERTopic for topic modeling and Sentence-BERT for sentence similarity. By incorporating LLMs like GPT and T5 for summarization and post-editing, PROMPTHEUS ensures that the generated summaries are accurate and coherent.

\section{PROMPTHEUS: A Framework for AI-Driven SLRs}

Despite significant advancements in AI-assisted SLRs, challenges remain in ensuring automated systems' accuracy, scalability, and relevance. Our proposed framework, PROMPTHEUS, introduces an integrated and fully automated SLR framework that enhances SLRs' Selection, Extraction, and Synthesis phases while maintaining human oversight during the Planning phase. PROMPTHEUS leverages advanced NLP techniques such as BERTopic for topic modeling and Sentence-BERT for sentence similarity to improve the precision and relevance of selected studies. Our system also integrates LLMs like GPT and T5 for summarization and post-editing, ensuring the generated summaries are accurate and coherent.

By introducing early-stage inclusion and exclusion criteria, PROMPTHEUS improves the rigor of study selection and reduces the likelihood of including irrelevant papers. This approach addresses the shortcomings identified by \citet{o2019question} and \citet{delaTorreLpez2023}, who emphasized the importance of integrating AI tools that improve efficiency without compromising the accuracy and comprehensiveness of systematic reviews, and also the challenges highlighted by \citet{affengruber2024rapid}, and \citet{Shaheen2023} who stressed the importance of balancing efficiency with comprehensive, high-quality reviews.

\subsection{General Overview}

\begin{figure}[!h]
    \resizebox{\columnwidth}{!}{
    \includegraphics{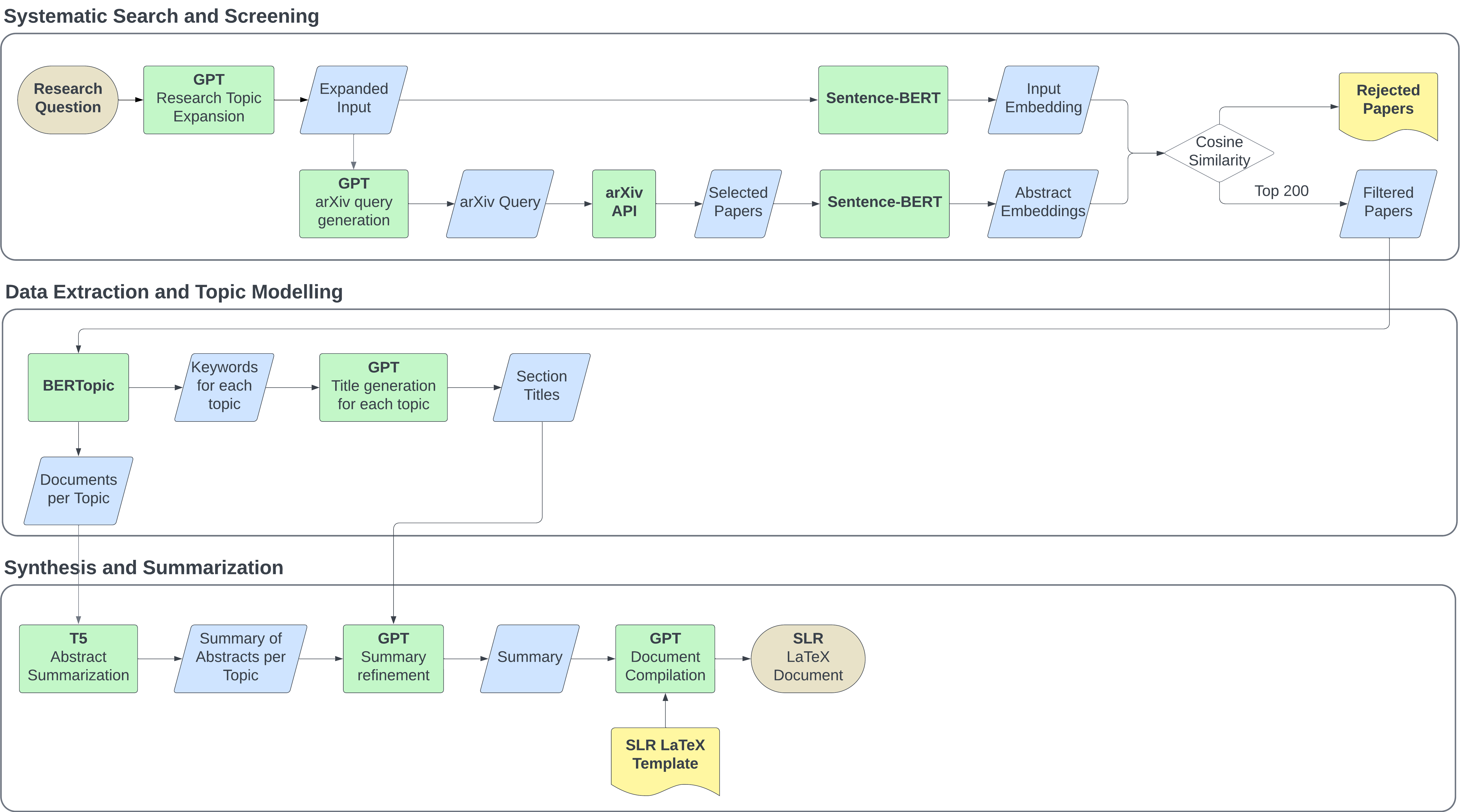}
    }
    \caption{The PROMPTHEUS framework consists of three phases: (1) Systematic Search and Screening using GPT and Sentence-BERT for paper selection, (2) Data Extraction and Topic Modeling with BERTopic and GPT for organizing and generating section titles, and (3) Synthesis and Summarization with T5 and GPT to refine and compile the findings into an SLR LaTeX document. This framework leverages NLP techniques and LLMs for an efficient and scalable SLR process.}
    \label{fig:promptheus}
\end{figure}

Our automated SLR pipeline architecture is organized into three interconnected phases: (1) Systematic Search and Screening, which identifies and selects relevant academic papers; (2) Data Extraction and Topic Modeling, which categorizes and organizes the selected studies; and (3) Synthesis and Summarization, which generates coherent summaries and integrates the findings into a structured review document. Each module employs specialized NLP techniques and LLMs, producing an efficient and scalable SLR process. Figure~\ref{fig:promptheus} presents the overall process.\\ 

\subsection{Systematic Search and Screening Module}

The Systematic Search and Screening Module is the foundation of the automated Systematic Literature Review process, which automates retrieving and filtering academic papers based on a user-defined research question or topic. This module addresses the limitations of traditional literature search methods, which often require extensive manual effort, by using LLMs and advanced NLP techniques to enhance the efficiency and precision of the search process. \\

\noindent
\textbf{Research Topic Expansion.}\\
The module begins with the user providing a research question or topic as input. The system leverages an LLM (GPT-3.5, GPT-4 or GPT-4o) to expand the initial input into a more detailed and semantically rich set of keywords and phrases to ensure the search captures a comprehensive range of relevant studies. This expansion is guided by a carefully crafted prompt that instructs the model to retain the core focus of the research topic while adding appropriate keywords and terms to cover variations and related concepts. Part of the prompt used for this task is:

\begin{tcolorbox}[colback=gray!10, colframe=black, width=\textwidth, sharp corners, boxrule=0.5mm]
\begin{quote}
\textit{System: "You are a knowledgeable AI specializing in generating expanded titles for research topics.  Your expanded titles should be concise and focus on capturing the core semantic meaning of a topic, suitable for creating informative embeddings for tasks like similarity comparisons."}\\

\noindent
\textit{User: "Task: Generate a slightly expanded title for the following research topic, keeping the core focus while potentially adding 1-2 highly relevant terms for improved semantic representation.\\
Topic: {title}\\
Guidelines:\\
                * Include essential keywords directly related to the topic.\\
                * If necessary, add 1-2 closely related terms to capture topic variations.\\
                * Avoid introducing new concepts or significantly altering the original title's meaning.\\
                * Keep the expanded title concise and focused on the core meaning.\\
Output format:\\
                * Provide the expanded title only. Do not include any additional explanations or commentary."
}
\end{quote}
\end{tcolorbox}

For instance, given the input topic "AI-based literature review", the LLM might generate an expanded version such as "AI-based literature review, automated systematic reviews, natural language processing for academic research synthesis". This expanded set of keywords ensures that the subsequent search covers a broader scope, capturing essential variations and closely related studies that might be overlooked.\\

\noindent
\textbf{Automated Query Generation.}\\
After expanding the research topic, the system constructs a structured search query tailored to the arXiv repository. This step is guided by another prompt instructing the LLM to craft a precise and targeted search query incorporating all relevant keywords and phrases. The model is asked to include fields such as title and abstract to refine the search further, ensuring that the retrieved literature aligns closely with the expanded topic. The prompt used for generating the search query is:

\begin{tcolorbox}[colback=gray!10, colframe=black, width=\textwidth, sharp corners, boxrule=0.5mm]
\begin{quote}
\textit{System: "You are a skilled research assistant specializing in crafting precise and effective search queries for the arXiv scientific paper repository."}\\

\noindent
\textit{User: "Task: Craft an effective search query tailored for the arXiv database, specifically designed to retrieve research papers on the following topic:\\
Topic: '{expanded\_title}'\\
 Guidelines:\\
                1.  Concise \& Precise: The query should be succinct yet accurately represent the core concept of the topic.\\
                2.  Key Terms: Incorporate the most relevant keywords or phrases directly associated with the topic.\\
                3.  Synonyms \& Variants (Optional): If applicable, include synonyms or alternative terms to broaden the search scope and capture nuanced variations of the topic.\\
                4.  Specificity:  Prioritize terms specific to the field or subfield to minimize irrelevant results.\\
                5.  arXiv Compatibility: Utilize operators like `ti:` (title) and `abs:` (abstract) to target specific fields within the arXiv entries.\\
Output format:\\
                * Provide the ArXiv query only. Do not include any additional explanations or commentary."}
\end{quote}
\end{tcolorbox}

The output of this prompt might produce a query such as \textit{(ti:"AI-based literature review" OR abs:("AI-based literature review" OR "automated systematic reviews")) AND (ti:"NLP" OR abs:"NLP")}. This structured query is then used to search the arXiv database through its API, retrieving up to 3000 academic papers that match the specified criteria. Once the search results are obtained, the module pre-processes the retrieved papers by extracting essential details such as paper ID, title, and abstract. The text is cleaned to ensure consistency and readability by removing unnecessary symbols and normalizing the format. This clean text is then used in the next stage of the module, where relevance filtering is performed.\\

\noindent
\textbf{Relevance Filtering Using Sentence Similarity.}
The module employs a similarity-based mechanism using Sentence-BERT embeddings to filter the most pertinent papers from the initial search results. It computes vector embeddings for both the expanded research topic and the cleaned abstracts of the retrieved papers. The cosine similarity between these embeddings is then calculated to assess the relevance of each paper. The top 200 papers with the highest similarity scores are selected for further analysis, ensuring the final literature set is focused and comprehensive. This structured approach significantly reduces manual effort while improving the quality and relevance of the selected studies, providing a robust foundation for subsequent stages.

Once the relevant papers are identified through systematic search and screening, the next step is to organize these documents into coherent themes using the Data Extraction and Topic Modeling Module.

\subsection{Data Extraction and Topic Modeling Module}

The Data Extraction and Topic Modeling Module automates organizing and categorizing selected academic papers into meaningful topics based on semantic content. The module leverages topic modeling and language generation techniques to create a structured literature representation, making identifying key research themes and subtopics easier. The module's core components include \textit{Topic Modeling and Document Clustering}, \textit{Keyword Extraction and Title Generation}, and \textit{Topic Report Generation}.\\

\noindent
\textbf{Topic Modeling and Document Clustering.}\\
Once the most relevant documents are selected from the initial screening phase, this module initiates by creating embeddings for the textual content of each document using a Sentence-BERT model. These embeddings capture the semantic information of the documents, allowing for an effective clustering of papers based on their conceptual similarities. Topic modeling uses the BERTopic algorithm, which groups documents into coherent clusters reflecting the selected literature's primary themes. The number of topics and the minimum topic size are dynamically adjusted based on the size and content of the dataset to ensure that the generated topics are both meaningful and interpretable.\\

\noindent
\textbf{Keyword Extraction and Title Generation.}\\
After clustering the documents into distinct topics, the system extracts keywords for each topic, summarising the main themes in that cluster. The keywords are input into a language model, such as GPT-3.5, GPT-4, or GPT-4o, to generate concise and descriptive titles for each topic. This process is guided by a structured prompt instructing the language model to create topic titles that accurately represent the essence of the keywords while maintaining clarity and relevance. The prompt used for this task is:

\begin{tcolorbox}[colback=gray!10, colframe=black, width=\textwidth, sharp corners, boxrule=0.5mm]
\begin{quote}
\textit{System: "You are an experienced researcher specializing in literature reviews. You are adept at crafting concise, informative, and engaging topic names for subsections that accurately reflect the content and guide the reader."}\\

\noindent
\textit{User: "Task: Create a clear and concise topic name for a subsection in a literature review. The subsection covers the following keywords: {topic\_keywords}:"} \\
\textit{Guidelines:}\\
\textit{* Length: Aim for 1-5 words.}\\
\textit{* Accuracy: Ensure the topic name precisely reflects the keywords' meaning.}\\
\textit{* Relevance: The name should fit within the broader context of a literature review.}\\
\textit{* Informativeness: Clearly indicate the subsection's focus to the reader.}\\
\textit{* Engagement: Make the topic name interesting and inviting to read.}\\
\textit{Optional: If the keywords are too broad or ambiguous, suggest a more specific or narrowed-down focus within the topic.}\\
\textit{Output format:}\\
\textit{ * Provide the topic title only. Do not include any additional explanations or commentary.}
\end{quote}
\end{tcolorbox}

For instance, if the extracted keywords for a topic are "deep learning, neural networks, image recognition," the generated title might be “Deep Learning for Image Recognition.” This descriptive title provides an overview of the underlying theme of the clustered documents, making it easier for researchers to navigate through the literature.\\

\noindent
\textbf{Topic Report Generation.}\\
After generating the titles, the system compiles a comprehensive report that includes the list of documents under each topic, the topic keywords, and the generated titles. This hierarchical organization of literature enhances the comprehensiveness and accessibility of the review, as it delineates different research themes and subtopics, making it easier for researchers to identify key trends and gaps in the literature. The module's process is further supported by a series of iterations and parameter adjustments to refine the topic modeling. If the initial number of topics is too few or too many, the system dynamically tunes the parameters, such as the number of topics or the minimum size of a topic, to achieve optimal clustering. 

Overall, this module significantly enhances the efficiency and effectiveness of the systematic literature review process by automating the categorization of papers and generating meaningful insights into the core themes of the literature. It automates the categorization of documents, offering researchers valuable insights into the core themes of the literature and simplifying the identification of key trends and gaps.

\subsection{Synthesis and Summarization Module}

The Synthesis and Summarization Module generates concise and coherent summaries for each identified topic cluster, significantly reducing the manual effort typically required in literature review processes. This module utilizes transformer-based models, such as T5, to summarize abstracts and GPT-based models for post-editing, ensuring that the resulting content is well-structured and easy to understand.\\

\noindent
\textbf{Abstract Summarization with T5.}\\
The process begins by generating summaries for individual abstracts within each topic cluster using a transformer-based model like T5. This model is specifically configured to produce short yet comprehensive summaries that capture each document's key contributions and findings. The generated summaries retain essential details while significantly reducing the length of the original abstracts, making it easier to synthesize large volumes of research.\\

\noindent
\textbf{Topic-Level Summarization and Aggregation.}\\
After individual summaries are generated, they are aggregated into a comprehensive summary for each identified topic. This step synthesizes the insights from multiple papers within the same topic, offering a holistic view of the research contributions, trends, and open questions. The aggregated summaries provide a structured narrative highlighting the most significant findings across multiple studies.\\

\noindent
\textbf{Post-Editing and Refinement with GPT.}\\
To enhance the clarity, coherence, and flow of the aggregated summaries, a GPT-based model is employed for post-editing. The refinement process involves using a predefined prompt instructing GPT to improve readability and structure while preserving critical information. This step ensures that the final summaries are well-organized and suitable for inclusion in a structured literature review document. The following prompt is used for post-editing:

\begin{tcolorbox}[colback=gray!10, colframe=black, width=\textwidth, sharp corners, boxrule=0.5mm]
\begin{quote}
\textit{System: "You are an expert researcher specializing in literature reviews in the field of {title}. Your task is to meticulously refine and enhance machine-generated summaries of multiple research papers."}\\
\textit{User: Refine the following machine-generated summary for the section "{section\_name}" in a literature review titled "{title}"}\\
\textit{The original summary is a compilation of various papers. Please focus on retaining the most relevant information for this literature review section.}\\
\textit{Crucially, ensure the inclusion of in-text citations (e.g., \\citep{{kadir2024revealing}}) for all information directly sourced from the referenced documents. Feel free to shorten the section summary if it enhances clarity and conciseness, but prioritize keeping essential details and all relevant citations.}\\
\textit{Original Summary: {summary}}\\
\textit{Output format:}\\
\textit{* Provide only the revised summary. Do not include any additional explanations or commentary.}
\end{quote}
\end{tcolorbox}

This refinement results in a more precise and cohesive summary that better communicates the core literature of the topic.

\noindent
\textbf{Document Compilation and Report Generation.}\\
The final step is compiling the generated summaries and topics into a coherent literature review document. This module integrates all the synthesized content into a structured LaTeX document, which includes an introduction, background information, detailed literature synthesis for each topic, and a conclusion. The system also generates a BibTeX file with the references for all included papers, ensuring proper citation and academic integrity.

The document generation process uses GPT, ensuring the final output is professionally formatted and adheres to the desired layout and style. The module supports various formats for exporting the final report, including LaTeX and PDF, providing researchers with a polished, ready-to-use literature review.




\section{Experimental Setup}

The proposed automated SLR framework was evaluated using a comprehensive experimental setup to assess its performance across different stages of the review process. We used five distinct research topics for the experiments: "Explainable Artificial Intelligence (XAI)," "Virtual Reality (VR)," "Blockchain," "Large Language Models (LLMs)," and "Neural Machine Translation (NMT)." Each experiment focused on a specific phase of the proposed SLR framework: Systematic Search and Screening, Data Extraction and Topic Modeling, and Synthesis and Summarization.\\

\noindent
\textbf{Datasets.} We conducted experiments using five different research topics, each representing a unique area of academic research: Explainable Artificial Intelligence, Virtual Reality, Blockchain, Large Language Models, and
Neural Machine Translation. We collected the papers for each research topic from the arXiv database. We retrieved papers based on search queries generated by GPT-3.5 and GPT-4o models, with a maximum limit of $3000$ papers per query.\\

\noindent
\textbf{Experiments.} We designed four experiments to assess the system's performance across different phases: Systematic Search and Screening, Data Extraction and Topic Modeling, Synthesis and Summarization, and Document Compilation and Report Generation. We reported the results using various metrics, including topic coherence, ROUGE scores, readability scores, and cosine similarity.\\

\noindent
\textbf{Readability Analysis.} We evaluated the readability of the generated summaries and final LaTeX documents using the Flesch Reading Ease Score (FRES). The Flesch Reading Ease Score \cite{kincaid1975derivation} provides insight into how easily a text can be read and understood. Higher FRES scores indicate simpler reading material, while lower scores denote more complex and challenging passages. We computed FRES at different stages of the summarization and document generation process to assess how readability changes as the content is processed through T5 summarization, GPT post-editing, and final document generation.\\

\noindent
\textbf{Metrics.} To evaluate the quality and robustness of the proposed framework, we used the following metrics:

\begin{itemize}
    \item \textbf{Topic coherence.} Measures the semantic similarity between words in a topic, indicating how well the generated topics represent coherent and interpretable concepts. A higher coherence score suggests that the words within each topic are more closely related, making the topics more useful and understandable for further analysis \citep{coherencemetrics}.

    \item \textbf{ROUGE} stands for Recall-Oriented Understudy for Gisting Evaluation. It compares an automatically produced summary or translation against a set of reference summaries (typically human-produced). ROUGE evaluates various aspects, such as the overlap of n-grams, word sequences, and word pairs between the machine-generated output and the reference.

    \item \textbf{ROUGE-1} measures the overlap of unigrams (single words) between the generated and reference abstracts. ROUGE-1 is particularly useful for evaluating summarization techniques because it captures the essential content and ensures that key information from the original text is retained in the summary.

    \item \textbf{Precision for ROUGE-1} measures the fraction of relevant instances among the retrieved cases, indicating how much of the generated summary is present in the reference text, which is the abstract in our case.

    \item \textbf{Recall for ROUGE-1} measures the fraction of instances retrieved over the total number of cases in the reference, indicating how much of the reference abstract is covered by the generated summary.

    \item \textbf{F1-Score for ROUGE-1} is the harmonic mean of precision and recall, providing a balance between the two metrics.

    \item \textbf{Cosine Similarity} measures the similarity between two non-zero vectors of an inner product space, effectively capturing the semantic closeness between the generated text and the reference text. Cosine similarity was used to evaluate the semantic alignment of abstracts with expanded topics during the Systematic Search and Screening phase.

    \item \textbf{Flesch Reading Ease Score (FRES)} \citep{kincaid1975derivation} provides insight into how easily a piece of text can be read and understood. The FRES formula considers sentence length and syllable count, with higher scores indicating simpler and more accessible text. We computed the FRES for three stages: T5-generated summaries, GPT post-edited sections, and the final LaTeX document.  The formula is as follows:

    \begin{equation}
    \text{FRES} = 206.835 - 1.015 \left( \frac{\text{total words}}{\text{total sentences}} \right) - 84.6 \left( \frac{\text{total syllables}}{\text{total words}} \right)
    \end{equation}

    This formula provides a measure of how easy a text is to read. Higher scores indicate easier-to-read material, while lower scores denote more difficult passages. 

    \item \textbf{Number of Papers Retrieved} indicates the coverage of the search query and its ability to find relevant literature.

    \item \textbf{Number of Papers Filtered} reflects the number of papers that passed an initial relevance filter based on the research topic.

    \item \textbf{Total CPU Time} is the computational time required for generating queries, retrieving papers, and filtering results.

\end{itemize}

\noindent
\textbf{Hardware.} Experiments were conducted on a Google Colab environment using an Intel Xeon CPU @ 2.20GHz (2 cores, 56MB cache), with 12.7 GB of RAM and 107.7 GB of disk space.

\subsection{Experiment 1: Systematic Search and Screening}

This experiment evaluated the effectiveness of GPT-3.5 and GPT-4o in generating queries for retrieving research papers from the arXiv repository. Given their capabilities in generating structured and contextually rich queries, we sought to compare the two models regarding their retrieval performance, efficiency, and computational cost. The experiment aimed to identify which model performs better across various research topics.

We selected five diverse research topics for this evaluation: Explainable Artificial Intelligence, Virtual Reality, Blockchain, Large Language Models, and Neural Machine Translation. For each topic, we measured three key performance indicators: Number of Papers Retrieved, Number of Papers Filtered, and CPU Time.\\

\noindent
\textbf{Results and Analysis.}\\
Results are summarized in Table \ref{tab:input_analysis}. GPT-4o consistently retrieved more papers than GPT-3.5 across all topics, indicating that GPT-4o generates comprehensive and relevant queries more effectively. For instance, GPT-4o retrieved 2833 papers for "Virtual Reality" compared to 1986 papers retrieved by GPT-3.5. Similarly, for "Explainable Artificial Intelligence," GPT-4o retrieved 1712 papers, surpassing the 1287 papers retrieved by GPT-3.5.

\begin{table}[!b]
\caption{Comparison of GPT-3.5 and GPT-4o in finding papers for the SLR process. CPU time indicates the total time for the entire automated SLR process.}
\label{tab:input_analysis}
\small
\begin{tabular}{@{}llcc@{}}
\toprule
\textbf{Input} & \textbf{Model} & \textbf{CPU Time (s)} & \textbf{Papers Found} \\ 
\midrule
\multirow{2}{*}{Explainable Artificial Intelligence} & GPT-3.5 & 1213 & 1287 \\
                                                     & GPT-4o  & 1555 & 1712 \\
\multirow{2}{*}{Virtual Reality}                     & GPT-3.5 & 1319 & 1986 \\
                                                     & GPT-4o  & 1496 & 2833 \\
\multirow{2}{*}{Blockchain}                          & GPT-3.5 & 1476 & 3000 \\
                                                     & GPT-4o  & 1563 & 3000 \\
\multirow{2}{*}{Large Language Models}               & GPT-3.5 & 1505 & 1400 \\
                                                     & GPT-4o  & 2115 & 3000 \\
\multirow{2}{*}{Neural Machine Translation}          & GPT-3.5 & 1648 & 2018 \\
                                                     & GPT-4o  & 1673 & 2073 \\
\bottomrule
\end{tabular}
\end{table}

While GPT-4o demonstrated superior retrieval capability, it also required significantly more computational time than GPT-3.5. For instance, the "Explainable Artificial Intelligence" topic took 1555 seconds to process using GPT-4o, whereas GPT-3.5 completed the same task in 1213 seconds—a difference of nearly 6 minutes. Similarly, GPT-4o required 2115 seconds to process the "Large Language Models" topic, which is approximately 10 minutes longer than GPT-3.5.

These results suggest that GPT-4o is more effective at generating queries that yield a more extensive set of relevant papers, making it well-suited for scenarios where comprehensive literature coverage is a priority. However, this increased retrieval capability comes at the cost of longer computational time, making GPT-4o less ideal for scenarios where efficiency and speed are critical considerations.

In conclusion, GPT-4o is preferable for use cases prioritizing comprehensive retrieval over computational efficiency, while GPT-3.5 may be better for time-sensitive applications. This insight provides a basis for selecting the appropriate LLM based on the specific requirements of different phases in the systematic literature review process.

\subsection{Experiment 2: Data Extraction and Topic Modelling}

This experiment evaluated the quality of topics generated during the data extraction and topic modeling phase. The goal was to determine how well the BERTopic algorithm organized the retrieved literature into meaningful and coherent themes.

We used the topic coherence metric from Gensim to measure the quality of the generated topics. Topic coherence quantifies the semantic similarity between words within a topic, indicating how well the topics represent coherent and interpretable concepts. This measure has been validated as a reliable method for assessing topic models in previous work by \citet{TopicEval2015}. Their study evaluated over 237,912 coherence measures across six benchmark datasets and demonstrated that specific combinations of coherence metrics correlate highly with human ratings, setting a standard for evaluating topic models.

Our experiment applied Gensim’s implementation of the coherence metric to assess the topics generated from documents retrieved using GPT-3.5 and GPT-4o queries. This metric ensures that the topics produced are statistically sound and interpretable to human evaluators.\\

\noindent
\textbf{Results and Analysis.}\\
This experiment assessed the semantic coherence of topics generated from the documents retrieved using GPT-3.5 and GPT-4o queries. As shown in Table \ref{tab:topic_analysis}, coherence scores for most topics fall between 0.4 and 0.5, indicating a moderate level of topic quality. This range suggests that the topics are generally coherent and interpretable but could be further refined.

\begin{table}[!b]
\caption{Topic coherence analysis using Gensim's topic coherence metric for GPT-3.5 and GPT-4o generated queries}
\label{tab:topic_analysis}
\small
\begin{tabular}{@{}llcc@{}}
\toprule
\textbf{Input} & \textbf{Model} & \textbf{Topic Coherence} & \textbf{Number of Topics} \\ 
\midrule
\multirow{2}{*}{Explainable Artificial Intelligence} & GPT-3.5 & 0.467 & 5 \\
                                                     & GPT-4o  & 0.422 & 6 \\
\multirow{2}{*}{Virtual Reality}                     & GPT-3.5 & 0.434 & 8 \\
                                                     & GPT-4o  & 0.481 & 8 \\
\multirow{2}{*}{Blockchain}                          & GPT-3.5 & 0.411 & 8 \\
                                                     & GPT-4o  & 0.475 & 5 \\
\multirow{2}{*}{Large Language Models}               & GPT-3.5 & 0.428 & 5 \\
                                                     & GPT-4o  & 0.473 & 5 \\
\multirow{2}{*}{Neural Machine Translation}          & GPT-3.5 & 0.477 & 6 \\
                                                     & GPT-4o  & 0.470 & 7 \\
\bottomrule
\end{tabular}
\end{table}

For instance, the topic coherence score for "Explainable Artificial Intelligence" was 0.467 using GPT-3.5 queries and 0.422 using GPT-4o queries, indicating that both models produce moderately coherent topics. Similarly, for "Virtual Reality," GPT-4o achieved a coherence score of 0.481 compared to 0.434 by GPT-3.5, showing that GPT-4o produced slightly better-organized topics for this research area.

Although these scores indicate that the generated topics are generally coherent, they are lower than previous benchmarks, such as the BERTopic model achieving scores of 0.681 and 0.432 on different datasets as reported by \citet{coherencemetrics}. This suggests that while our system can generate meaningful topics, there is potential for further improvements in topic coherence to match or exceed these higher benchmark scores.

\subsection{Experiment 3: Synthesis and Summarization}

This experiment assessed the performance of the Synthesis and Summarization phase of our automated literature review framework. We evaluated the quality of the generated summaries using ROUGE scores to determine their relevance and content retention. Additionally, the readability of each summary was analyzed using the Flesch Reading Ease metric. The primary goal was to determine how effectively the system condenses and synthesizes information from multiple research papers while maintaining coherence and relevance.\\

\noindent
\textbf{ROUGE Score Analysis.}\\
We used the ROUGE-1 metric to compare the content overlap between the machine-generated summaries and the abstracts of the selected research papers, which served as reference texts. ROUGE-1 measures the degree of overlap in unigrams (single words) between the generated summaries and reference texts, making it suitable for evaluating content retention and relevance.

The evaluation was conducted in three stages:
\begin{itemize}
    \item[] \textbf{Abstract Generated Summaries using T5}: These serve as the baseline summaries generated by the T5 model, which captures the core content of the abstracts.

    \item[] \textbf{Post-Edited Generate Summaries using GPT}: GPT-based models refine these summaries to enhance readability, coherence, and overall structure.

    \item[] \textbf{Document Compilation and Report Generation using LaTeX}: Comprehensive sections formatted as LaTeX documents that integrate information from multiple summaries, providing a cohesive and structured literature overview.
\end{itemize}

We computed ROUGE-1 precision, recall, and F1 scores for each stage. While all three metrics provide valuable insights, we focused primarily on precision. High precision indicates that the summaries retain the most pertinent information from the reference abstracts, minimizing irrelevant details.\\

\begin{table}[!h]
\centering
\small
\caption{ROUGE-1 Scores for the T5-generated Summaries (P = Precision, R = Recall, F1 = F-measure), with the selected abstracts as reference}
\label{tab:rouge_scores_summarized}
\begin{tabular}{@{}llccc@{}}
\toprule
\multirow{2}{*}{\textbf{Input}} & \multirow{2}{*}{\textbf{Model}} & \multicolumn{3}{c}{\textbf{ROUGE}} \\ 
\cmidrule(lr){3-5}
 &  & \textbf{P} & \textbf{R} & \textbf{F1} \\ 
\midrule
\multirow{2}{*}{Explainable Artificial Intelligence} & GPT-3.5 & 0.963 & 0.405 & 0.570 \\
                                                     & GPT-4o  & 0.964 & 0.387 & 0.552 \\
\multirow{2}{*}{Virtual Reality}                     & GPT-3.5 & 0.967 & 0.418 & 0.583 \\
                                                     & GPT-4o  & 0.969 & 0.425 & 0.591 \\
\multirow{2}{*}{Blockchain}                          & GPT-3.5 & 0.967 & 0.401 & 0.567 \\
                                                     & GPT-4o  & 0.968 & 0.400 & 0.567 \\
\multirow{2}{*}{Large Language Models}               & GPT-3.5 & 0.966 & 0.376 & 0.540 \\
                                                     & GPT-4o  & 0.965 & 0.381 & 0.546 \\
\multirow{2}{*}{Neural Machine Translation}          & GPT-3.5 & 0.965 & 0.462 & 0.625 \\
                                                     & GPT-4o  & 0.965 & 0.460 & 0.623 \\
\bottomrule
\end{tabular}
\end{table}

The results in Table \ref{tab:rouge_scores_summarized} indicate that both GPT-3.5 and GPT-4o models achieved high precision (P) scores across all inputs, demonstrating that the generated summaries contain a significant proportion of relevant content present in the reference abstracts. However, the recall scores are relatively lower, reflecting that not all content from the reference abstracts is captured in the summaries. This is expected since we want to capture only the relevant information in the abstracts. In this table, there is no significant benefit in choosing GPT-4o instead of GPT-3.5, as the T5 model computes the summary. GPT, at this stage, was only used to gather documents by creating the ArXiv query.\\

\noindent
\textbf{Post-Editing Stage Results}

The post-editing phase is crucial in refining the machine-generated summaries produced by the T5 model. This stage utilizes GPT-based models to enhance the initial summaries' clarity, coherence, and structure. The objective is to condense and reorganize the content while preserving the most relevant information. Post-editing is essential for transforming raw summaries into well-structured sections that align with the broader context of an SLR.

Table \ref{tab:rouge_scores_post_edited} presents the ROUGE-1 scores for the summaries after being refined by GPT-based models. 

\begin{table}[!h]
\centering
\small
\caption{ROUGE-1 Scores for GPT Post-Edited Sections (P = Precision, R = Recall, F1 = F-measure)}
\label{tab:rouge_scores_post_edited}
\begin{tabular}{@{}llccc@{}}
\toprule
\multirow{2}{*}{\textbf{Input}} & \multirow{2}{*}{\textbf{Model}} & \multicolumn{3}{c}{\textbf{ROUGE}} \\
\cmidrule(lr){3-5}
 &  & \textbf{P} & \textbf{R} & \textbf{F1} \\
\midrule
\multirow{2}{*}{Explainable Artificial Intelligence} & GPT-3.5 & 0.922 & 0.029 & 0.055 \\
                                                     & GPT-4o  & 0.884 & 0.063 & 0.118 \\
\multirow{2}{*}{Virtual Reality}                     & GPT-3.5 & 0.920 & 0.029 & 0.057 \\
                                                     & GPT-4o  & 0.906 & 0.063 & 0.118 \\
\multirow{2}{*}{Blockchain}                          & GPT-3.5 & 0.924 & 0.038 & 0.072 \\
                                                     & GPT-4o  & 0.897 & 0.029 & 0.056 \\
\multirow{2}{*}{Large Language Models}               & GPT-3.5 & 0.919 & 0.028 & 0.055 \\
                                                     & GPT-4o  & 0.901 & 0.042 & 0.080 \\
\multirow{2}{*}{Neural Machine Translation}          & GPT-3.5 & 0.911 & 0.034 & 0.066 \\
                                                     & GPT-4o  & 0.883 & 0.075 & 0.138 \\
\bottomrule
\end{tabular}
\end{table}

The results show a clear drop in recall after post-editing compared to the initial T5-generated summaries, reflecting the focus on refining and condensing content. This reduction is expected as the goal is to produce well-structured, concise sections for the systematic literature review (SLR). Despite the decrease in recall, precision scores remain high, ensuring the retained information is relevant and concise.

GPT-4o demonstrates higher recall than GPT-3.5, indicating its ability to retain more content during post-editing. However, the F1 scores, which balance precision and recall, show only a slight advantage for GPT-4o, suggesting that both models perform similarly in maintaining a good balance between content relevance and retention.

Post-editing with GPT significantly improved the precision of the summaries, ensuring that the most relevant information was retained, even though recall slightly decreased. GPT-4o showed a slight edge in content retention, making it more suitable for comprehensive literature reviews\\

\noindent
\textbf{Final LaTeX Document Evaluation}\\
The final LaTeX documents generated by the system were evaluated to understand their effectiveness in creating cohesive literature review sections that integrate information from multiple sources. 

As presented in Table \ref{tab:rouge_scores_latex_doc}, both GPT-3.5 and GPT-4o achieved exceptionally high precision scores (e.g., 0.991 for GPT-3.5 and 0.989 for GPT-4o in the "Explainable Artificial Intelligence" topic), indicating that the final documents are highly aligned with the reference abstracts in terms of relevance. However, the recall scores for these documents were relatively low, which is expected given that the final LaTeX documents are designed to provide comprehensive literature review sections, not direct summaries of the abstracts. These documents incorporate additional background information, contextual insights, and synthesized content from various sources, which broadens the scope and naturally reduces recall scores.

\begin{table}[!h]\centering
\small
\caption{ROUGE-1 Scores for GPT-generated Final LaTeX Document (P = Precision, R = Recall, F1 = F-measure)}
\label{tab:rouge_scores_latex_doc}
\begin{tabular}{@{}llccc@{}}
\toprule
\multirow{2}{*}{\textbf{Input}} & \multirow{2}{*}{\textbf{Model}} & \multicolumn{3}{c}{\textbf{ROUGE}} \\
\cmidrule(lr){3-5}
 &  & \textbf{P} & \textbf{R} & \textbf{F1} \\
\midrule
\multirow{2}{*}{Explainable Artificial Intelligence} & GPT-3.5 & 0.991 & 0.018 & 0.036 \\
                                                     & GPT-4o  & 0.989 & 0.026 & 0.049 \\
\multirow{2}{*}{Virtual Reality}                     & GPT-3.5 & 0.988 & 0.015 & 0.029 \\
                                                     & GPT-4o  & 0.972 & 0.034 & 0.065 \\
\multirow{2}{*}{Blockchain}                          & GPT-3.5 & 0.987 & 0.023 & 0.045 \\
                                                     & GPT-4o  & 0.995 & 0.028 & 0.054 \\
\multirow{2}{*}{Large Language Models}               & GPT-3.5 & 0.995 & 0.021 & 0.042 \\
                                                     & GPT-4o  & 0.990 & 0.039 & 0.076 \\
\multirow{2}{*}{Neural Machine Translation}          & GPT-3.5 & 0.976 & 0.021 & 0.041 \\
                                                     & GPT-4o  & 0.969 & 0.038 & 0.074 \\
\bottomrule
\end{tabular}
\end{table}

Despite this, the F1 scores, which balance precision and recall, show that the final documents maintain a strong balance between relevance and content coverage. GPT-4o generally achieved higher recall scores than GPT-3.5, suggesting it is more effective at incorporating additional relevant content while maintaining overall coherence. This makes GPT-4o particularly useful in scenarios where comprehensive literature coverage is essential.

While GPT-4o offers advantages in retaining more comprehensive content, GPT-3.5 remains a competitive option for generating concise and highly relevant summaries. Future efforts could focus on improving recall without sacrificing precision, allowing for even more comprehensive and well-rounded literature review sections.

\subsection{Experiment 4: Readability Score}

We used the Flesch Reading Ease Score to evaluate the readability of the generated summaries and final documents at different stages of the document generation process. This metric provides insights into how accessible the text is to a general audience, with higher scores indicating easier-to-read content. Readability was evaluated for the T5-generated summaries, GPT post-edited summaries, and the final LaTeX documents.

Table \ref{table:flesch_reading_ease} outlines the interpretation of Flesch Reading Ease scores, with lower scores indicating text that requires a higher level of education to comprehend. In our evaluation, we compared these scores across the stages of the automated SLR process to assess how the readability evolved from the initial summarization to the final document creation.

\begin{table}[t]
\centering
\caption{Interpretation of Flesch Reading Ease Scores \citep{kincaid1975derivation,Flesch_Kincaid_readability_tests}}
\label{table:flesch_reading_ease}
\begin{tabular}{@{}ccl@{}}
\toprule
\textbf{Score} & \textbf{School Level (US)} & \textbf{Notes} \\
\midrule
100.00--90.00  & 5th grade                  & Very easy to read. Easily understood by an average 11-year-old student. \\
90.00--80.00   & 6th grade                  & Easy to read. Conversational English for consumers. \\
80.00--70.00   & 7th grade                  & Fairly easy to read. \\
70.00--60.00   & 8th \& 9th grade           & Plain English. Easily understood by 13- to 15-year-old students. \\
60.00--50.00   & 10th to 12th grade         & Fairly difficult to read. \\
50.00--30.00   & College                    & Difficult to read. \\
30.00--10.00   & College graduate           & Very difficult to read. Best understood by university graduates. \\
10.00--0.00    & Professional               & Extremely difficult to read. Best understood by university graduates. \\
\bottomrule
\end{tabular}
\end{table}

Table \ref{tab:readability_scores} presents the Flesch Reading Ease scores for each stage of the document generation process. These scores provide an overview of how readability changes as the content is transformed from initial summaries to refined, structured documents.

\begin{table}[b]
\caption{Readability Scores for T5-generated Summaries, GPT Post-Edited Sections, and GPT-generated Final LaTeX Document}
\label{tab:readability_scores}
\small
\centering
\begin{tabular}{@{}llcccc@{}}
\toprule
\textbf{Input} & \textbf{Model} & \textbf{T5Sum} & \textbf{GPTSec} & \textbf{GPTSLR} & \textbf{Baseline} \\
\midrule
\multirow{2}{*}{Explainable Artificial Intelligence} & GPT-3.5 & 8.450 & 30.942 & 29.945 & 50.827 \\
                                                     & GPT-4o  & 9.196 & 28.053 & 34.542 & \\
\multirow{2}{*}{Virtual Reality}                     & GPT-3.5 & 16.126 & 33.974 & 26.926 & 54.297 \\
                                                     & GPT-4o  & 14.481 & 32.560 & 26.237 & \\
\multirow{2}{*}{Blockchain}                          & GPT-3.5 & 6.874 & 22.158 & 19.024 & 47.461 \\
                                                     & GPT-4o  & 9.954 & 24.339 & 20.314 & \\
\multirow{2}{*}{Large Language Models}               & GPT-3.5 & 14.399 & 34.505 & 35.041 & 53.834 \\
                                                     & GPT-4o  & 21.734 & 30.466 & 33.798 & \\
\multirow{2}{*}{Neural Machine Translation}          & GPT-3.5 & 20.815 & 32.191 & 29.242 & 59.556 \\
                                                     & GPT-4o  & 22.508 & 27.106 & 31.754 & \\
\bottomrule
\end{tabular}
\end{table}

\noindent
\textbf{T5-Generated Summaries} exhibit low readability scores, indicating that the content is quite challenging to read. This outcome is expected due to the highly condensed nature of the T5-generated summaries, which prioritize brevity over readability, often lacking the narrative structure required for easier comprehension.

\noindent
The readability of the summaries significantly improves after the \textbf{GPT Post-Edited Sections}. The post-editing process refines the content by enhancing clarity, improving sentence structure, and providing a more coherent flow. This results in more accessible and readable sections, reflected in the enhanced Flesch scores.

\noindent
\textbf{Final LaTeX Generated Documents} show further improvements in readability. The additional structuring, formatting, and content synthesis contribute to easier-to-read documents than the earlier stages. However, while the readability has improved, it remains lower than the baseline.

\noindent
\textbf{Baseline Summaries exhibit the highest readability scores}, demonstrating that maintaining readability while summarizing and synthesizing content remains a challenge. The baseline scores highlight the gap between the generated summaries and the clarity of the original abstracts.

\noindent
The Flesch Reading Ease Scores improved progressively from T5-generated summaries to GPT post-edited sections and final LaTeX documents. However, despite these improvements, the readability of the generated documents remains lower than that of the original abstracts. This outcome underscores the challenge of maintaining high readability while compressing and synthesizing content, particularly in automated systems.

\subsection{Experiment 5: Sentence Similarity}
To further evaluate the effectiveness of our automated systematic literature review (SLR) system, we computed cosine similarity scores at various stages of document generation. This analysis quantifies how closely the generated summaries and final documents align with the original input queries, providing a measure of content retention and relevance. For comparison, a baseline (Random) was included, representing a document generated with random words, to serve as a control.

Cosine similarity measures the cosine of the angle between two vectors in a multidimensional space—here, these vectors represent text embeddings derived from the documents. Higher cosine similarity scores indicate greater alignment between the generated texts and the original input queries.

\begin{table}[h]
\centering
\small
\caption{Cosine Similarity between the embedding of the Input and the embeddings of Generated Documents (Abs = Abstracts, T5Sum = T5-generated Summaries, GPTSec = GPT Post-Edited Sections, GPTSLR = GPT-generated Final LaTeX Document, Random = Document created with randomly generated words)}
\label{tab:text_similarity}
\begin{tabular}{llccccc}
\multirow{2}{*}{\textbf{Input}} & \multirow{2}{*}{\textbf{Model}} & \multicolumn{5}{c}{\textbf{Cosine Similarity}} \\
\cmidrule(lr){3-7}
 &  & \textbf{Abs} & \textbf{T5Sum} & \textbf{GPTSec} & \textbf{GPTSLR} & \textbf{Random} \\
\midrule
\multirow{2}{*}{Explainable Artificial Intelligence} & GPT-3.5 & 0.644 & 0.587 & 0.718 & 0.763 & 0.118 \\
                                                     & GPT-4o  & 0.446 & 0.539 & 0.687 & 0.753 & ~\\
\multirow{2}{*}{Virtual Reality}                     & GPT-3.5 & 0.482 & 0.539 & 0.597 & 0.552 & 0.082 \\
                                                     & GPT-4o  & 0.613 & 0.624 & 0.602 & 0.689 & ~ \\
\multirow{2}{*}{Blockchain}                          & GPT-3.5 & 0.583 & 0.511 & 0.665 & 0.681 & 0.071 \\
                                                     & GPT-4o  & 0.646 & 0.587 & 0.550 & 0.612 & ~ \\
\multirow{2}{*}{Large Language Models}               & GPT-3.5 & 0.551 & 0.446 & 0.598 & 0.610 & 0.127 \\
                                                     & GPT-4o  & 0.436 & 0.619 & 0.591 & 0.691 & ~ \\
\multirow{2}{*}{Neural Machine Translation}          & GPT-3.5 & 0.622 & 0.691 & 0.692 & 0.694 & 0.108 \\
                                                     & GPT-4o  & 0.622 & 0.664 & 0.669 & 0.743 & ~ \\
\bottomrule
\end{tabular}
\end{table}

Table \ref{tab:text_similarity} presents the cosine similarity scores for each stage of the document generation process, offering insights into how well the system preserves content relevance across different stages:\\

\noindent
\textbf{Abstract Filtering.} The filtered abstracts generated by GPT-3.5 and GPT-4o exhibit high cosine similarity scores, demonstrating that the initial search and screening phase effectively identifies documents closely related to the input queries. Both models perform well at this stage, confirming the robustness of the search process.\\

\noindent
\textbf{T5-Generated Summaries.} The cosine similarity scores decrease slightly for the T5-generated summaries, which is expected. Summarization inherently condenses content and may omit some details, leading to a lower similarity with the full abstracts. However, the core information relevant to the input query remains retained, ensuring that the generated summaries focus on the main topics.\\

\noindent
\textbf{GPT Post-Edited Summaries.} The cosine similarity scores increase after the post-editing process by GPT. This improvement suggests that the GPT-based post-editing refines the structure and readability and enhances alignment with the original input. The post-editing process ensures that the key content is retained while improving the coherence of the generated sections. GPT-4o generally outperforms GPT-3.5 in maintaining content similarity, indicating that GPT-4o is more effective at preserving relevant information.\\

\noindent
\textbf{Final LaTeX Documents.} The final documents generated by GPT continue to exhibit high similarity scores, indicating that the synthesis and summarization process effectively retains the relevance of the content. The structured nature of LaTeX documents ensures that core themes from the input queries are well-represented. GPT-4o again shows slightly better performance than GPT-3.5, further suggesting that it is better at understanding and incorporating relevant content throughout the document generation process.\\

\noindent
\textbf{Random Baseline:} As expected, the cosine similarity scores for the randomly generated document are very low. This serves as a control, validating the significance of the similarity scores observed for the generated summaries and final documents.\\

\noindent
Overall, the system demonstrates robustness in generating documents that remain closely aligned with the original input queries, ensuring that the synthesized literature reviews preserve essential information while improving readability and structure.

\subsection{Experiment 6: Finding the Optimal Number of Papers for SLR}

\begin{figure}[!h]
    \resizebox{\columnwidth}{!}{
    \includegraphics[]{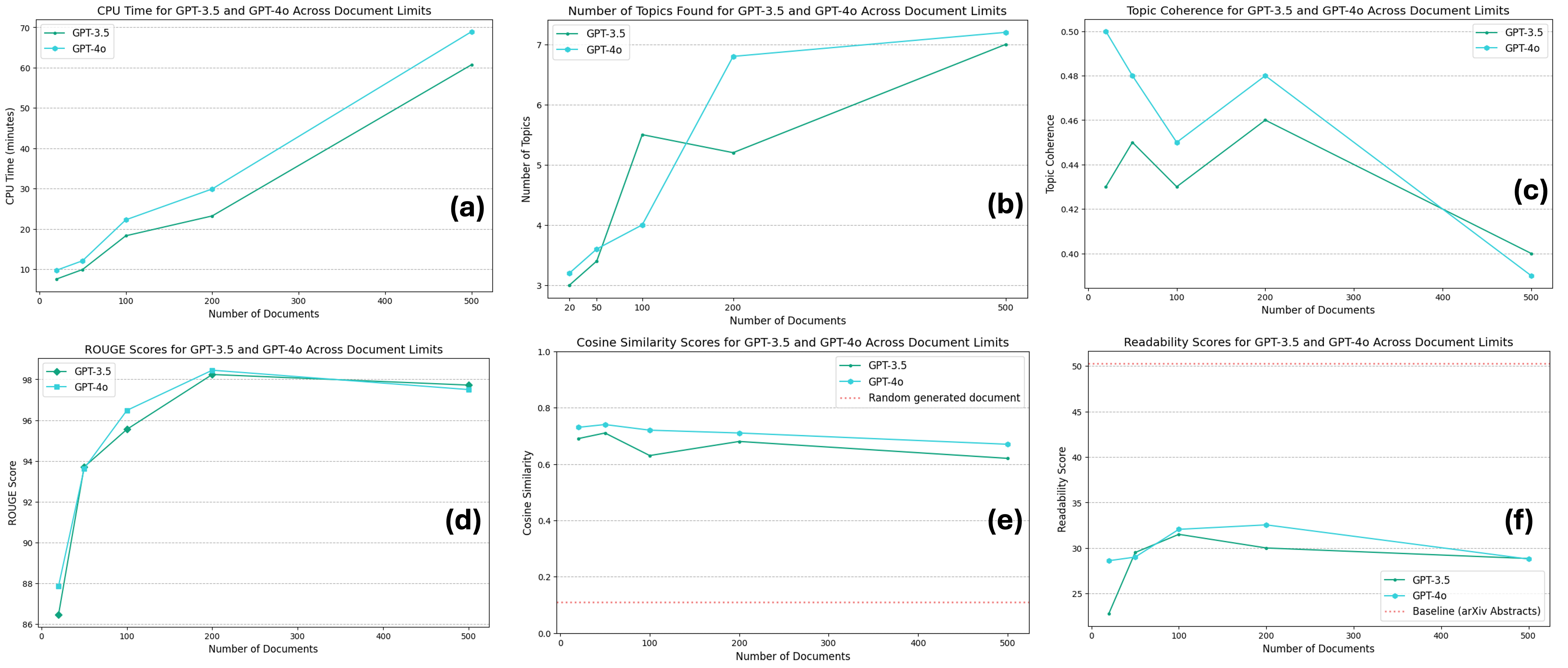}
    }
    \caption{Performance metrics across different document limits for GPT-3.5 and GPT-4o in the SLR process. (a) CPU Time: GPT-4o consistently requires more time than GPT-3.5 as the number of documents increases, reflecting its computational complexity. (b) Number of Topics: GPT-4o identifies more topics, indicating a finer level of clustering. (c) Topic Coherence: Coherence is stable up to 200 documents for both models, but it declines as more documents are added, suggesting overfitting or noise. (d) ROUGE Scores: Summarization quality improves and plateaus around 200 documents. (e) Cosine Similarity: Both models show stable alignment with input queries, with diminishing returns beyond 200 documents. (f) Readability Scores: Readability peaks around 200 documents before declining, suggesting this as the optimal limit for accessible summaries.}
    \label{fig:num_papers}
\end{figure}

This section explores several key performance metrics to determine the optimal number of papers to include in the SLR process. The metrics analyzed include CPU time, number of topics identified, topic coherence, ROUGE scores, readability scores, and cosine similarity scores. These metrics are used to assess the impact of different document limits on the quality and efficiency of the SLR. We recommend the most effective document limit that balances performance and computational resources based on the analysis results. Figure~\ref{fig:num_papers} presents the results obtained.\\

\noindent
\textbf{CPU Time.} Figure~\ref{fig:num_papers}-(a) shows the CPU time shows how computational requirements scale with the number of documents processed. As the document count increases, CPU time rises significantly, with GPT-4o consistently requiring more time than GPT-3.5. This indicates that although GPT-4o can potentially offer more accurate results, it demands more computational resources, which is a trade-off to consider when processing large volumes of documents.

\noindent
\textbf{Number of Topics Found.} In Figure~\ref{fig:num_papers}-(b), BERTopic identifies an increasing number of topics as more documents are processed. GPT-4o consistently identifies more topics than GPT-3.5 across all document limits. This suggests that GPT-4o is more adept at detailed clustering, potentially offering a more nuanced breakdown of the literature. However, after a certain threshold, the increase in topics may not necessarily translate to better quality but rather more fragmented groupings.\\

\noindent
\textbf{Topic Coherence.} The Topic Coherence metric measures the semantic similarity within the topics identified, providing insight into the quality of the generated clusters. Figure~\ref{fig:num_papers}-(c) illustrates the quality of the topics generated based on the semantic similarity of words within them. GPT-3.5 and GPT-4o maintain relatively stable topic coherence scores of up to 200 documents. Beyond this point, coherence begins to drop slightly for both models, likely due to overfitting or noise introduced by an excessive number of documents. This reinforces that 200 documents strike an optimal balance between quality and quantity regarding topic coherence.\\

\noindent
\textbf{ROUGE Scores for Summarization Quality.} The ROUGE scores measure how well the generated summaries align with reference abstracts, focusing on content retention. Figure~\ref{fig:num_papers}-(d) shows that as the number of documents increases, the ROUGE scores improve, peaking around 200. This suggests that the system becomes better at generating summaries that capture the core content of the papers as more documents are processed. However, beyond the 200-document threshold, the improvement in ROUGE scores plateaus, indicating that additional documents do not contribute significantly to better summarization. This implies that while increasing the document count improves the system’s ability to summarize effectively, there is little benefit to going beyond 200 documents regarding content retention and quality.\\

\noindent
\textbf{Cosine Similarity for Content Alignment.} The Cosine Similarity scores measure how closely the generated documents align with the input queries, indicating relevance and focus. Figure~\ref{fig:num_papers}-(e) shows that GPT-3.5 and GPT-4o achieve high similarity scores across all document limits, stabilizing around 200 documents. This indicates that 200 documents provide sufficient information to produce outputs well-aligned with the original research query without overwhelming the system with excess data. The plateau in similarity scores beyond this threshold suggests that additional documents do not significantly enhance the relevance of the generated summaries. Therefore, 200 documents appear to be the most efficient choice for maintaining high alignment with the research objectives while minimizing computational overhead.\\

\noindent
\textbf{Readability Scores.} We use the Flesch Reading Ease \citep{kincaid1975derivation} metric to evaluate how accessible and easy to read the generated summaries are. Figure~\ref{fig:num_papers}-(f) indicates that the readability scores increase as more documents are processed, reaching their highest point, around 200. This suggests that the generated summaries become clearer and easier to read as the system processes more documents, possibly due to having a more comprehensive pool of content to draw from. However, readability scores decline slightly after 200 documents, indicating that the system might introduce more complex or fragmented language as the document count grows. This highlights that 200 documents offer the best balance for generating summaries that are both informative and easy to read.\\

\noindent
\textbf{Optimal Number of Papers }
Based on the analysis of the above metrics, 200 documents emerge as the optimal document limit for the SLR process. At this threshold, the system provides high-quality summaries, maintains strong topic coherence, and produces readable and relevant outputs without excessive computational resources. Using over 200 documents leads to diminishing returns, particularly regarding topic coherence, readability, and cosine similarity. Thus, we recommend 200 documents as the ideal balance between performance and efficiency for conducting automated systematic literature reviews.

\section{Discussion}

The results presented in this study demonstrate the potential of the proposed automated SLR framework to streamline and enhance the process of conducting literature reviews. By integrating advanced NLP techniques and LLMs such as GPT-3.5 and GPT-4o, the framework automates systematic search, data extraction, topic modeling, and summarization stages. However, a critical analysis of the results reveals both strengths and areas for improvement.\\

\noindent
\textbf{GPT-4o retrieves more papers than GPT-3.5}. The experiments revealed that GPT-4o consistently outperformed GPT-3.5 in retrieving a larger number of papers across all research topics. This suggests that GPT-4o is better at generating more comprehensive and contextually rich search queries. The ability of GPT-4o to retrieve more papers is beneficial in scenarios where exhaustive literature coverage is important, such as systematic reviews and meta-analyses, as it ensures that a wider array of relevant research is considered. However, this improved retrieval capacity may also introduce more irrelevant or low-quality papers, necessitating more robust filtering mechanisms.\\

\noindent
\textbf{High ROUGE-1 precision scores for both models.} Both GPT-3.5 and GPT-4o demonstrated high ROUGE-1 precision scores during the summarization phase, indicating that the generated summaries retained a significant amount of relevant content from the reference abstracts. This suggests that the models effectively focus on the most important information when creating summaries, which is critical in systematic reviews where maintaining the relevance of summarized content is paramount. However, the relatively lower recall scores reflect that some content was omitted during summarization, which may be intentional to avoid overwhelming the reader with excessive detail. The high precision with lower recall suggests a bias toward conciseness, which can be advantageous in certain contexts but may require adjustment depending on the goals of the review.\\

\noindent
\textbf{Post-editing improved precision but reduced recall.} The post-editing phase significantly improved the precision of the generated summaries but reduced recall, indicating that while the content became more concise and focused, some relevant details were omitted. This aligns with the goal of post-editing, which is to refine and streamline the summaries for clarity and coherence. GPT-4o demonstrated higher recall than GPT-3.5 in this phase, suggesting it more effectively retained content during post-editing. This slight advantage highlights GPT-4o's ability to balance relevance and conciseness better, making it more suitable for generating comprehensive yet readable summaries in systematic reviews.\\

\noindent
\textbf{GPT-4o achieved higher recall in final LaTeX documents.} The final LaTeX documents generated by GPT-4o achieved higher recall scores than those generated by GPT-3.5, indicating that GPT-4o was more successful in incorporating additional relevant content while maintaining coherence. This makes GPT-4o particularly advantageous for use cases that require comprehensive literature coverage, as the final documents generated by GPT-4o were better at synthesizing information from multiple sources. However, this increase in recall may come at the cost of readability, as the additional content could make the final documents more complex and challenging to navigate. Future work could explore optimizing the balance between recall and readability in final document generation.\\

\noindent
\textbf{Readability improved through post-editing and final document generation.} The Flesch Reading Ease Scores demonstrated a clear improvement in readability from the initial T5-generated summaries to the GPT post-edited sections and final LaTeX documents. This suggests that the post-editing process significantly enhanced the clarity and coherence of the summaries, making them easier to read and understand. However, despite these improvements, the readability of the generated documents remained lower than the baseline abstracts. This outcome reflects the inherent difficulty in maintaining high readability while condensing and synthesizing technical content. Future work could explore more advanced techniques to improve readability, especially in the post-editing phase, to close the gap with the original abstracts.\\

\noindent
\textbf{Cosine similarity confirms robust content retention.} Cosine similarity scores across all stages of document generation were high, confirming that the system retained key content from the original input queries. The post-editing and final document generation stages further improved content alignment, particularly with GPT-4o, which generally outperformed GPT-3.5 in maintaining content relevance. These results suggest that both models effectively ensure the generated summaries and documents stay focused on the core topics of the input queries, making them reliable tools for systematic literature reviews. The consistently high similarity scores also validate the robustness of retrieval, summarization, and synthesis, ensuring that essential information is not lost throughout the stages.

The findings from this study underscore the utility of combining GPT models and NLP techniques to automate key phases of systematic literature reviews, from retrieval to summarization. While GPT-4o demonstrates superior performance in content retrieval and recall, GPT-3.5 remains competitive for tasks prioritizing efficiency and conciseness. The framework shows promise in automating extensive literature reviews with relatively high precision and robust content retention. However, challenges remain, particularly in optimizing readability and balancing recall with document complexity. Future work should focus on refining the post-editing processes, improving the coherence and accessibility of generated documents, and ensuring that the system remains adaptable to diverse academic domains. Enhancing the framework's ability to filter irrelevant or lower-quality content will strengthen its applicability in high-demand, resource-intensive reviews.

\section{Ethical Considerations and Limitations}

A key limitation of this study is that it only analyzed proprietary models, specifically OpenAI's GPT-3.5 and GPT-4o, and did not include open-source models like LLaMA or Falcon. This is important because open-source models are becoming increasingly popular for research and practical applications due to their accessibility and customization potential. By focusing only on proprietary models, this study misses the opportunity to evaluate the performance, bias mitigation strategies, and transparency advantages that open-source models may offer.

\noindent
\textbf{Bias in Literature Selection.} AI models can introduce bias into the literature selection process. These models are trained on large, potentially unbalanced datasets, which can skew the selection towards more popular or well-represented topics, ignoring less-covered research areas. Future work should explore ways to address this bias, possibly through fairness-aware algorithms or more inclusive data sources.

\noindent
\textbf{Hallucination and Misinformation.} Both GPT-3.5 and GPT-4o are prone to generating content that is not directly grounded in the input data, which can lead to inaccurate summaries. This is particularly risky in a systematic literature review where factual accuracy is critical. Adding validation steps, such as human review or factual grounding mechanisms, would help mitigate this issue.

\noindent
\textbf{Exclusion of Proprietary Models.} By not including freely available and modifiable models like LLaMA or Falcon, this study does not address how open-source solutions could improve transparency, reproducibility, and control in the SLR process. These models offer the potential for better alignment with specific research needs and ethical considerations such as data privacy and bias control.

\noindent
\textbf{Scalability.} Although the system performed well with up to 3000 papers per query, larger datasets may introduce computational challenges. Future iterations should improve scalability, possibly by adopting distributed computing methods.

Future studies should incorporate open-source models to compare their effectiveness and address the broader needs of the academic community, offering more flexibility, transparency, and cost control.

\section{Conclusions}

This study proposed PROMPTHEUS, an automated SLR framework that integrates advanced NLP techniques and large language models to streamline the literature review process. By automating systematic search, data extraction, topic modeling, and synthesis, the framework effectively manages the growing volume of academic literature. Our experiments across five research topics demonstrated the system’s strengths in selecting relevant papers, retaining key content, and improving readability in post-editing stages. However, areas like recall and topic coherence still require improvement.

Despite achieving high precision, lower recall scores, and moderate topic coherence, these metrics suggest that some relevant content was omitted and topics could be better organized. GPT-4o outperformed GPT-3.5 in recall and content retention. However, readability, though improved in post-editing, remained below the clarity of the original abstracts.

In conclusion, PROMPTHEUS contributes to automating systematic literature reviews by combining advanced NLP techniques and large language models. Our framework addresses the growing challenges of managing vast academic literature by streamlining critical processes while maintaining high precision and content relevance. However, the trade-offs between precision, recall, and readability underscore further refinement, particularly in improving topic coherence and ensuring that the most relevant content is consistently included. Looking ahead, the future potential of PROMPTHEUS is vast. We will continue to optimize these elements while incorporating open-source models to improve flexibility, transparency, and scalability, empowering the academic community to more effectively manage the ever-rapidly-growing body of research with enhanced comprehensiveness and efficiency.

\section{Acknowledgments}

This work was supported by the UNESCO Chair on AI \& VR 

\section{Authors' Contributions}

Joao Pedro Fernandes Torres: methodology, investigation, software development, evaluation, document writing, document revision.\\
Catherine Mulligan: supervision, investigation, document revision.\\
Joaquim Jorge: supervision, investigation, document writing, document revision.\\
Catarina Moreira: supervision, investigation, methodology, evaluation, document writing, document revision.

\section{Funding}

This work was funded through Fundação para a Ciência e Tecnologia with references 2022.09212.PTDC and  DOI:10.54499/UIDB/50021/2020, DOI:10.54499/DL57/2016/CP1368/CT0002

\section{Data Availability}

The PROMPTHEUS framework is open-source and available at \url{https://github.com/joaopftorres/PROMPTHEUS.git}. Our repository also includes the outputs generated by PROMPTHEUS for the five research queries analyzed in this paper: Explainable Artificial Intelligence, Virtual Reality, Blockchain, Large Language Models, and Neural Machine Translation. The datasets used are public and obtained from \url{https://arxiv.org/}.

\section{Declarations}

\subsection{Ethics approval and consent to participate}

Not applicable

\subsection{Consent for publication}

Not applicable

\subsection{Competing interests}

The authors declare that they have no competing interests.

\bibliography{sn-article}%

\begin{thebibliography}{34}
\providecommand{\natexlab}[1]{#1}
\providecommand{\url}[1]{{#1}}
\providecommand{\urlprefix}{URL }
\providecommand{\doi}[1]{\url{https://doi.org/#1}}
\providecommand{\eprint}[2][]{\url{#2}}
 \bibcommenthead

\bibitem[{Affengruber et~al(2024{\natexlab{a}})Affengruber, van~der Maten, Spiero, Nussbaumer-Streit, Mahmi{\'c}-Kaknjo, Ellen, Goossen, Kantorova, Hooft, Riva et~al}]{affengruber2024exploration}
Affengruber L, van~der Maten MM, Spiero I, et~al (2024{\natexlab{a}}) An exploration of available methods and tools to improve the efficiency of systematic review production: a scoping review. BMC Medical Research Methodology 24(1):210

\bibitem[{Affengruber et~al(2024{\natexlab{b}})Affengruber, Nussbaumer-Streit, Hamel, Van~der Maten, Thomas, Mavergames, Spijker, and Gartlehner}]{affengruber2024rapid}
Affengruber L, Nussbaumer-Streit B, Hamel C, et~al (2024{\natexlab{b}}) Rapid review methods series: Guidance on the use of supportive software. BMJ evidence-based medicine

\bibitem[{Bolanos et~al(2024)Bolanos, Salatino, Osborne, and Motta}]{bolanos2024artificial}
Bolanos F, Salatino A, Osborne F, et~al (2024) Artificial intelligence for literature reviews: Opportunities and challenges. \urlprefix\url{https://arxiv.org/abs/2402.08565}, \eprint{2402.08565}

\bibitem[{Brown et~al(2020)Brown, Mann, Ryder, Subbiah, Kaplan, Dhariwal, Neelakantan, Shyam, Sastry, Askell, Agarwal, Herbert-Voss, Krueger, Henighan, Child, Ramesh, Ziegler, Wu, Winter, Hesse, Chen, Sigler, Litwin, Gray, Chess, Clark, Berner, McCandlish, Radford, Sutskever, and Amodei}]{gpt3}
Brown TB, Mann B, Ryder N, et~al (2020) Language models are few-shot learners. \eprint{2005.14165}

\bibitem[{Chappell et~al(2023)Chappell, Edwards, Watkins, Marshall, and Graziadio}]{chappell2023machine}
Chappell M, Edwards M, Watkins D, et~al (2023) Machine learning for accelerating screening in evidence reviews. Cochrane Evidence Synthesis and Methods 1(5):e12021

\bibitem[{Chu and Evans(2021)}]{CanonicalProgress}
Chu JSG, Evans JA (2021) Slowed canonical progress in large fields of science. Proceedings of the National Academy of Sciences 118(41). \doi{10.1073/pnas.2021636118}, \urlprefix\url{https://www.pnas.org/content/118/41/e2021636118}, {\href{https://arxiv.org/abs/https://www.pnas.org/content/118/41/e2021636118.full.pdf}{{https://www.pnas.org/content/118/41/e2021636118.full.pdf}}}

\bibitem[{Cierco~Jimenez et~al(2022)Cierco~Jimenez, Lee, Rosillo, Cordova, Cree, Gonzalez, and Indave~Ruiz}]{cierco2022machine}
Cierco~Jimenez R, Lee T, Rosillo N, et~al (2022) Machine learning computational tools to assist the performance of systematic reviews: A mapping review. BMC Medical Research Methodology 22(1):322

\bibitem[{Dennst{\"a}dt et~al(2024)Dennst{\"a}dt, Zink, Putora, Hastings, and Cihoric}]{dennstadt2024title}
Dennst{\"a}dt F, Zink J, Putora PM, et~al (2024) Title and abstract screening for literature reviews using large language models: an exploratory study in the biomedical domain. Systematic Reviews 13(1):158

\bibitem[{Gates et~al(2020)Gates, Gates, DaRosa, Elliott, Pillay, Rahman, Vandermeer, and Hartling}]{gates2020decoding}
Gates A, Gates M, DaRosa D, et~al (2020) Decoding semi-automated title-abstract screening: findings from a convenience sample of reviews. Systematic reviews 9:1--12

\bibitem[{Guo et~al(2024)Guo, Jiang, Zhao, Long, Feng, Gu, Xu, Li, Huang, and Du}]{guo2024rapid}
Guo Q, Jiang G, Zhao Q, et~al (2024) Rapid review: A review of methods and recommendations based on current evidence. Journal of Evidence-Based Medicine

\bibitem[{Hamel et~al(2021)Hamel, Hersi, Kelly, Tricco, Straus, Wells, Pham, and Hutton}]{hamel2021guidance}
Hamel C, Hersi M, Kelly SE, et~al (2021) Guidance for using artificial intelligence for title and abstract screening while conducting knowledge syntheses. BMC Medical Research Methodology 21:1--12

\bibitem[{Kharawala et~al(2021)Kharawala, Mahajan, and Gandhi}]{kharawala2021artificial}
Kharawala S, Mahajan A, Gandhi P (2021) Artificial intelligence in systematic literature reviews: a case for cautious optimism. Journal of Clinical Epidemiology 138:243

\bibitem[{Kincaid(1975)}]{kincaid1975derivation}
Kincaid J (1975) Derivation of new readability formulas (automated readability index, fog count and flesch reading ease formula) for navy enlisted personnel. Chief of Naval Technical Training

\bibitem[{Li et~al(2024)Li, Sun, and Tan}]{li2024evaluating}
Li M, Sun J, Tan X (2024) Evaluating the effectiveness of large language models in abstract screening: a comparative analysis. Systematic Reviews 13(1):219

\bibitem[{Masoumi et~al(2024)Masoumi, Amirkhani, Sadeghian, and Shahraz}]{masoumi2024natural}
Masoumi S, Amirkhani H, Sadeghian N, et~al (2024) Natural language processing (nlp) to facilitate abstract review in medical research: the application of biobert to exploring the 20-year use of nlp in medical research. Systematic Reviews 13(1):107

\bibitem[{Moreno-Garcia et~al(2023)Moreno-Garcia, Jayne, Elyan, and Aceves-Martins}]{moreno2023novel}
Moreno-Garcia CF, Jayne C, Elyan E, et~al (2023) A novel application of machine learning and zero-shot classification methods for automated abstract screening in systematic reviews. Decision Analytics Journal 6:100162

\bibitem[{Ofori-Boateng et~al(2024)Ofori-Boateng, Aceves-Martins, Wiratunga, and Moreno-Garcia}]{ofori2024towards}
Ofori-Boateng R, Aceves-Martins M, Wiratunga N, et~al (2024) Towards the automation of systematic reviews using natural language processing, machine learning, and deep learning: a comprehensive review. Artificial intelligence review 57(8):200

\bibitem[{Okoli(2015)}]{SLRguide}
Okoli C (2015) A guide to conducting a standalone systematic literature review. Communications of the Association for Information Systems 37. \doi{10.17705/1CAIS.03743}

\bibitem[{O’Connor et~al(2019)O’Connor, Tsafnat, Thomas, Glasziou, Gilbert, and Hutton}]{o2019question}
O’Connor AM, Tsafnat G, Thomas J, et~al (2019) A question of trust: can we build an evidence base to gain trust in systematic review automation technologies? Systematic reviews 8:1--8

\bibitem[{Page et~al(2021)Page, McKenzie, Bossuyt, Boutron, Hoffmann, Mulrow, Shamseer, Tetzlaff, Akl, Brennan, Chou, Glanville, Grimshaw, Hróbjartsson, Lalu, Li, Loder, Mayo-Wilson, McDonald, McGuinness, Stewart, Thomas, Tricco, Welch, Whiting, and Moher}]{Page2021}
Page MJ, McKenzie JE, Bossuyt PM, et~al (2021) The prisma 2020 statement: an updated guideline for reporting systematic reviews. BMJ p n71. \doi{10.1136/bmj.n71}, \urlprefix\url{http://dx.doi.org/10.1136/bmj.n71}

\bibitem[{Perlman-Arrow et~al(2023)Perlman-Arrow, Loo, Bobrovitz, Yan, and Arora}]{perlman2023real}
Perlman-Arrow S, Loo N, Bobrovitz N, et~al (2023) A real-world evaluation of the implementation of nlp technology in abstract screening of a systematic review. Research Synthesis Methods 14(4):608--621

\bibitem[{Qureshi et~al(2023)Qureshi, Shaughnessy, Gill, Robinson, Li, and Agai}]{qureshi2023chatgpt}
Qureshi R, Shaughnessy D, Gill KA, et~al (2023) Are chatgpt and large language models “the answer” to bringing us closer to systematic review automation? Systematic Reviews 12(1):72

\bibitem[{Raffel et~al(2023)Raffel, Shazeer, Roberts, Lee, Narang, Matena, Zhou, Li, and Liu}]{t5}
Raffel C, Shazeer N, Roberts A, et~al (2023) Exploring the limits of transfer learning with a unified text-to-text transformer. \eprint{1910.10683}

\bibitem[{Rahimi et~al(2023)Rahimi, Hoover, Mimno, Naacke, Constantin, and Amann}]{coherencemetrics}
Rahimi H, Hoover JL, Mimno D, et~al (2023) Contextualized topic coherence metrics. \eprint{2305.14587}

\bibitem[{Robledo et~al(2023)Robledo, Grisales~Aguirre, Hughes, and Eggers}]{robledo2023vista}
Robledo S, Grisales~Aguirre AM, Hughes M, et~al (2023) “hasta la vista, baby”--will machine learning terminate human literature reviews in entrepreneurship? Journal of Small Business Management 61(3):1314--1343

\bibitem[{R{\"o}der et~al(2015)R{\"o}der, Both, and Hinneburg}]{TopicEval2015}
R{\"o}der M, Both A, Hinneburg A (2015) Exploring the space of topic coherence measures. In: Proceedings of the eight International Conference on Web Search and Data Mining, Shanghai, February 2-6, \urlprefix\url{http://svn.aksw.org/papers/2015/WSDM_Topic_Evaluation/public.pdf}

\bibitem[{Saeidmehr et~al(2024)Saeidmehr, Steel, and Samavati}]{Saeidmehr2024spiral}
Saeidmehr A, Steel PDG, Samavati FF (2024) Systematic review using a spiral approach with machine learning. Systematic Reviews 13(1):32. \doi{10.1186/s13643-023-02421-z}, \urlprefix\url{https://doi.org/10.1186/s13643-023-02421-z}

\bibitem[{Sami et~al(2024)Sami, Rasheed, Kemell, Waseem, Kilamo, Saari, Duc, Systä, and Abrahamsson}]{sami2024}
Sami AM, Rasheed Z, Kemell KK, et~al (2024) System for systematic literature review using multiple ai agents: Concept and an empirical evaluation. \urlprefix\url{https://arxiv.org/abs/2403.08399}, \eprint{2403.08399}

\bibitem[{Shaheen et~al(2023)Shaheen, Shaheen, Ramadan, Hefnawy, Ramadan, Ibrahim, Hassanein, Ashour, and Flouty}]{Shaheen2023}
Shaheen N, Shaheen A, Ramadan A, et~al (2023) Appraising systematic reviews: a comprehensive guide to ensuring validity and reliability. Frontiers in Research Metrics and Analytics 8. \doi{10.3389/frma.2023.1268045}, \urlprefix\url{http://dx.doi.org/10.3389/frma.2023.1268045}

\bibitem[{Speckemeier et~al(2022)Speckemeier, Niemann, Wasem, Buchberger, and Neusser}]{speckemeier2022methodological}
Speckemeier C, Niemann A, Wasem J, et~al (2022) Methodological guidance for rapid reviews in healthcare: a scoping review. Research synthesis methods 13(4):394--404

\bibitem[{de~la Torre-López et~al(2023)de~la Torre-López, Ramírez, and Romero}]{delaTorreLpez2023}
de~la Torre-López J, Ramírez A, Romero JR (2023) Artificial intelligence to automate the systematic review of scientific literature. Computing 105(10):2171–2194. \doi{10.1007/s00607-023-01181-x}, \urlprefix\url{http://dx.doi.org/10.1007/s00607-023-01181-x}

\bibitem[{T{\'o}th et~al(2024)T{\'o}th, Berek, Gul{\'a}csi, P{\'e}ntek, and Zrubka}]{toth2024automation}
T{\'o}th B, Berek L, Gul{\'a}csi L, et~al (2024) Automation of systematic reviews of biomedical literature: a scoping review of studies indexed in pubmed. Systematic Reviews 13(1):174

\bibitem[{Ware and Mabe(2015)}]{STMreport}
Ware M, Mabe M (2015) The stm report: An overview of scientific and scholarly journal publishing

\bibitem[{Wikipedia(2024)}]{Flesch_Kincaid_readability_tests}
Wikipedia (2024) {Flesch–Kincaid readability tests} --- {W}ikipedia{,} the free encyclopedia. \url{http://en.wikipedia.org/w/index.php?title=Flesch\%E2\%80\%93Kincaid\%20readability\%20tests&oldid=1214583693}

\end{thebibliography}

\end{document}